\documentclass[runningheads]{llncs}
\usepackage{graphicx}
\usepackage{amsmath,amssymb} % define this before the line numbering.
\usepackage{xcolor}
\usepackage[width=122mm,left=12mm,paperwidth=146mm,height=193mm,top=12mm,paperheight=217mm]{geometry}
\usepackage{adjustbox}
\usepackage{multirow}

\begin{document}

%\pagecolor{black!30}
% \linenumbers
\pagestyle{headings}
\mainmatter

\title{Improving Object Counting with \\Heatmap Regulation}

\titlerunning{Improving Object Counting with Heatmap Regulation}

\authorrunning{Aich and Stavness}

\author{Shubhra Aich and Ian Stavness}

\institute{Department of Computer Science, University of Saskatchewan, Canada\\
\email{\{s.aich, ian.stavness\}@usask.ca}}

\maketitle

\begin{abstract}
In this paper, we propose a simple and effective way to improve one-look regression models for object counting from images. We use class activation map visualizations to illustrate the drawbacks of learning a pure one-look regression model for a counting task. Based on these insights, we enhance one-look regression counting models by regulating activation maps from the final convolution layer of the network with coarse ground-truth activation maps generated from simple dot annotations. We call this strategy heatmap regulation (HR). We show that this simple enhancement effectively suppresses false detections generated by the corresponding one-look baseline model and also improves the performance in terms of false negatives. Evaluations are performed on four different counting datasets --- two for car counting (CARPK, PUCPR+), one for crowd counting (WorldExpo) and another for biological cell counting (VGG-Cells). Adding HR to a simple VGG front-end improves performance on all these benchmarks compared to a simple one-look baseline model and results in state-of-the-art performance for car counting.

\end{abstract}

\section{Introduction}
\label{intro}

Counting object instances from images and videos is a common and practical computer vision task found in a range of applications, such as counting vehicles from aerial images, crowd counting for surveillance, biological cell counting for medical diagnosis, plant counting for image based plant phenotyping, and so on. The problem of object counting can be considered a subproblem of object detection, which is in turn a subproblem of instance-level segmentation. In that sense, the instance segmentation pipeline is sufficient for the other two of its sub-domains --- object detection and counting \cite{mask-rcnn}. However, the development of supervised algorithms for instance segmentation and object detection demands ground-truth annotation with much higher specificity compared to mere object counting frameworks. Obtaining large-scale annotated datasets with fine granularity is a prohibitively time-consuming process. Thus, specialized counting approaches that use lighter weight image labels are worth pursuing for applications where an object count alone is needed.

\begin{figure}[t!]
\centering
\includegraphics[scale=0.26]{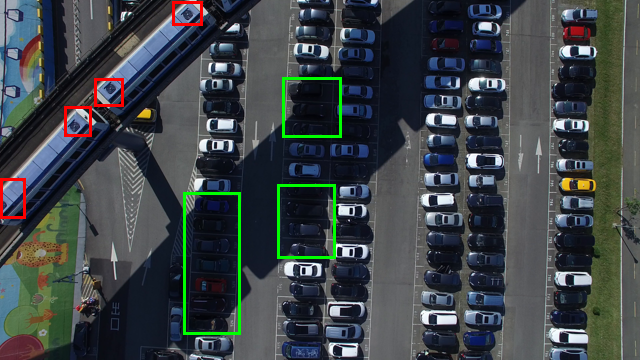}
\includegraphics[scale=0.26]{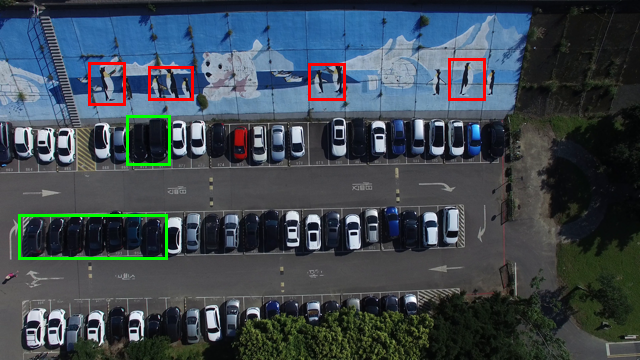} \\
\includegraphics[scale=0.26]{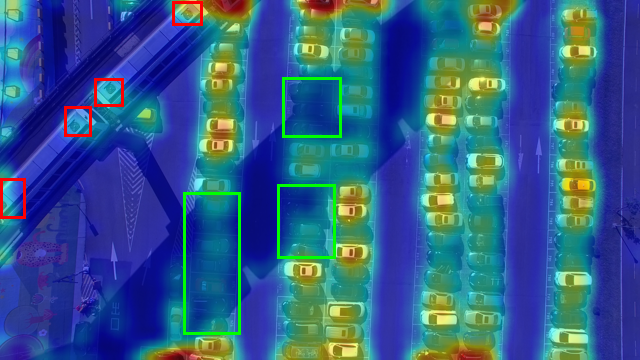}
\includegraphics[scale=0.26]{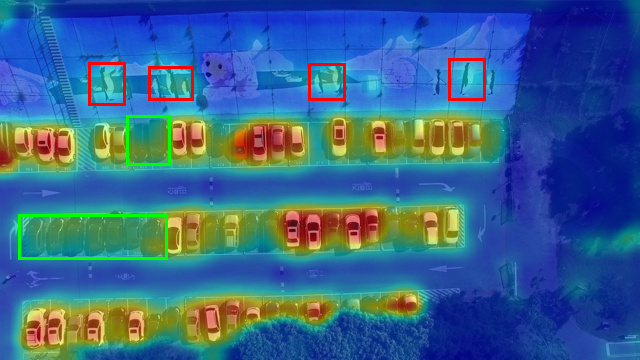} \\
\includegraphics[scale=0.26]{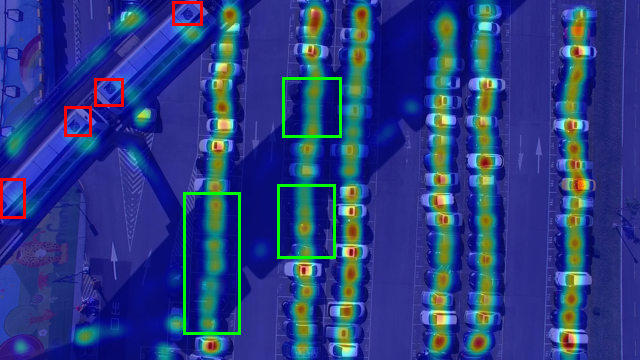}
\includegraphics[scale=0.26]{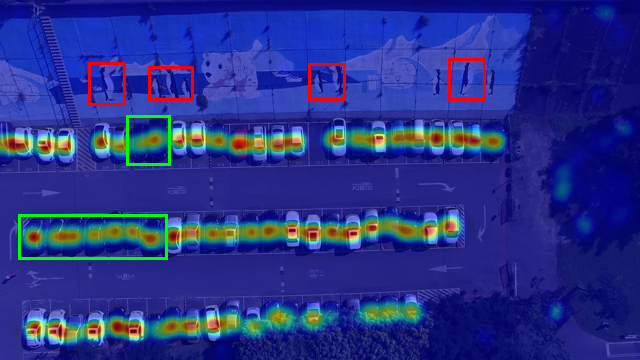}
\caption{Sample images from the from CARPK \cite{lpn-carpk} dataset (top row) along with superimposed CAM heatmaps generated by the baseline model (middle row) and the model enhanced with heatmap regulation (bottom row). The baseline model exhibits probable false detections for parts of the train and painted wall (red boxes) and probable missed instances for black-colored cars and regions in shadow (green boxes). Heatmap regulation fixes both of these cases and results in more compact activations.}
\label{fig:drawbacks}
\end{figure}

\
The predominant deep learning approach for object counting alone is one-look regression, where the model directly predicts a scalar count for an input image. These networks either adopt a classification architecture, where the number of possible output units is a slight overestimate predefined based on the training data \cite{cowc}, or have a single output unit generating a real numbered value as close to the target count as possible \cite{deepwheat,best-cvppp,aich-cvppp,dpp}. Both variants can be classified under the category of high dimensional nonlinear regression models, where the number of predictors is proportional to the number of input pixels and the number of response variables is either one for the single output case or the highest possible count for the classification case.

While no explicit localization information is provided to one-look regression models for counting, visualizations of activation heatmaps have suggested that such models focus, to a certain extent, on salient regions of the image~\cite{deepwheat,best-cvppp}. For most counting problems, object instances will share similar texture, color, and shape properties in the image space, hence the network will automatically learn to recognize most of them as part of the regression problem.
However, a potential pitfall of this counting approach is that a one-look model may miss a few harder-to-detect instances. In order to compensate for these missed detections, and produce the correct count label, the network would falsely mark a few background sub-regions, which have similar regional image properties, as possible object instances instead.

Indeed we observe this phenomena when applying a one-look regression model to a recent car counting dataset and visualizing with class activation maps \cite{cam-mit}: dark/shadowed cars are missed and background sub-regions are activated (Figure \ref{fig:drawbacks}). These visualizations demonstrate that posing the counting problem as a mere nonlinear regression problem without further constraints has contextual limitations and can cause an unavoidable generalization error. This is because the only constraint for the weakly supervised regression network is to map the input image into a count as close to the target as possible, without explicit information about the target object properties in the image. In other words, it has an abundance of unsupervised spatial context without any prior information about the kind of contextual relationship to be exploited for learning to count.
The problem of generating false positives because of missing true positives in one-look counting models can only be perceived, and thus avoided, by the network if some information is available about probable object locations in the image alongside the scalar target value and the loss function is formulated taking this fact into account.

In this paper, we propose a novel way to provide additional contextual information to simple one-look counting models through dot annotations that incur similar annotation effort as obtaining counts alone.
We generate a coarse ground-truth Gaussian activation map (GAM) or saliency map from dot annotations available for counting. Next, we incorporate the idea of back-propagating the differential error between the predicted class activation map (CAM) \cite{cam-mit} and our ground-truth GAM alongside the counting errors with the goal to suppress false detections and enhance false negatives. This error channel can be easily injected into light weight architectures with minimal computational expense. We call this additional error suppression strategy heatmap regulation (HR). To the best of our knowledge, although CAM is widely used to visualize the final saliency map of CNN architectures, the regulation of saliency maps via CAM for training the models is unique to our work. We evaluate the HR strategy on four different object counting datasets --- two for cars (CARPK, PUCPR+) \cite{lpn-carpk}, one for crowds \cite{cross-scene-sjtu} and the other for biological cells (VGG-Cells) \cite{zisserman-nips-count}. Adding HR to one-look models results in more accurate counts and more compact saliency maps in all cases. Our approach achieves state-of-the-art accuracy on both car datasets and obtains comparable performance on the rest with a simple VGG-GAP baseline model. As our HR approach is easy to implement and incurs little computational burden, we expect this idea to be a reliable enhancement for many one-look deep learning approaches to counting problems.

% ========================================================================== %
\section{Related Works}
\textbf{Density map estimation:} A number of previous works incorporate the concept of density map estimation into the counting pipeline. The most influential early work on counting via density map estimation is done by Lempitsky and Zisserman \cite{zisserman-nips-count}. They generate pixel-level ground-truth density map from the dot annotations using one Gaussian kernel per object instance followed by density map estimation via linear transformation of the pixel-level feature representations using regularized risk minimization. Their ground-truth density map generation process is similar to our ground-truth activation map generation (Gaussian kernels on dot annotations). However, we used simple per-pixel $L_1$ metric for the downscaled activation map in our paper which is not suitable for training the models in their framework because of the traditional feature extraction based approach. Extensions to this basic idea are provided in \cite{fiaschi2012,interactive-count,xie2016,arteta2016,segui2015}.

Fiaschi et al. \cite{fiaschi2012} replaces the linear model of Lempitsky and Zisserman \cite{zisserman-nips-count} with a regression random forest to predict the density map over the input as well as object count. The authors in \cite{interactive-count} enhance the approach of \cite{zisserman-nips-count} by providing options to interactively receive dot annotations from the users and iteratively correcting the annotations via further visualization.

Xie et al. \cite{xie2016} propose fully convolutional regression networks which can be trained on arbitrary input size to predict the final density map. Arteta et al. \cite{arteta2016} propose a multi-task deep architecture to predict the density map from noisy crowdsourcing annotations. The tasks include foreground segmentation from a less accurate ground-truth labels or tri-maps, prediction of the density map, and an additional uncertainty map prediction representing the variability among multiple annotations. Segui et al. \cite{segui2015} explore the strength of deep features and show the deep models learn multi-purpose feature representation even while being trained only on a particular task. For example, they show the network can learn to recognize digits while being trained for counting the number of even digits on a synthetic MNIST dataset. Encoder-decoder architecture is used in \cite{average-pridmore} to estimate density map followed by non-maximum suppression to estimate the number of wheat spikes and spikelets.

Alternative spatial map approximation approaches are also proposed in \cite{xie2015,average-pridmore}. Xie et al. \cite{xie2015} uses deep learning to predict proximity map instead of density or count map, where the output pixel values indicate the proximity from the nearest cell center. The Count-ception paper \cite{countception} replaces the density map estimation by count map estimation using a fully convolutional network \cite{fcn} with very small input size($32 \times 32$) and then use the idea of redundant spatial coverage to estimate the final count.

Our approach differs from all these density, count or proximity map estimation methods in the sense that our focus is not on estimating any of these maps, rather we emphasize on getting better counting performance by regulating the final activation map or CAM using the Gaussian activation map generated from the simple dot annotations.
\newline \newline
\noindent\textbf{Counting networks:} Convolutional networks have been tremendously successful in generating region proposals and bounding boxes with associated class probabilities for different categories of objects \cite{faster-rcnn,yolo}. The layout proposal network \cite{lpn-carpk} paper enhances the idea of bounding box generation with a domain-specific prior representing the spatial layout of objects. Also, instance-level segmentation is accomplished with substantial accuracy with state-of-the-art convolutional and recurrent network architectures and their hybrids \cite{mask-rcnn,ris,instance-seg-toronto}. Although object counts are readily available from these detection and segmentation frameworks, they need more detailed ground-truth annotations which are hard to acquire in a large-scale. Considering this fact, where counting is the only task at hand, convolutional networks are employed as a high-dimensional regression network to generate a real-valued or discrete count from the input image directly~\cite{acm-multimedia-2015,cowc,aich-cvppp,best-cvppp,deepwheat,dpp}. In this paper, we work on improving the design limitations of these one-look counting models without sacrificing their simplicity and efficiency.

Crowd counting approaches based on traditional computer vision algorithms mostly work on comparatively low-density crowds, whereas recent deep learning approaches exhibit a tremendous performance boost on high-density crowds. Sindagi and Patel \cite{survey-crowd} provide a comprehensive survey of these approaches. Here, we discuss the ones most relevant to ours. The cross-scene crowd counting paper \cite{cross-scene-sjtu} has a spirit similar to ours in the sense that they also compare their generated density maps with probable Gaussian maps using a deep architecture. The authors use a very small network (only 3 convolution and nonlinear activation layers) and input size ($72\times72$) followed by 3 fully connected (FC) layers. As a result, the network needs to switch between local$-$global$-$local contexts for semi-local density map approximation with the additional risk of overfitting. Furthermore, the process for generating Gaussian density maps for the crowd scenes are dependent on the perspective projection of the camera to take into account the perspective change of the length of the human body in different locations in the image, whereas our Gaussian map generation process is independent of the perspective of the instances. Finally, in the crowd counting paper, additional effort is needed by means of an extra FC layer to generate the density map and $2-\text{stage}$ training strategy optimizing the density map estimation loss followed by the counting one, whereas our CAM-GAM comparison performs simultaneous training for both losses.

\section{Our Approach}
\label{method}

Our motivating hypothesis is that although a simple one-look model naturally learns to localize most of the target object instances from the image only from scalar count information, the lack of explicit information about object properties or the regions in the image from where it should learn to count is ultimately determinate to the counting performance. We expect that the one-look model tries to identify sub-regions in the image with very similar properties, like texture, color, etc. up to the number equal to the target count. As a result, for most of the true instances, the model correctly localizes them, but fails for harder-to-detect instances or for instances for which the regional properties differ from most others. To compensate for the discrepancy in the count compared to the target count caused by these harder instances, the simple one-look model may falsely learn to detect a few background sub-regions as the true candidates which have similar properties as many of the true object instances in the training dataset. Consequently, the one-look regression model suffers from a comparatively higher generalization error, part of which is completely unavoidable due to the lack of ground-truth information about the object properties and locations. We evaluate this hypothesis by visualizing and regulating CAM as shown in Figure \ref{fig:drawbacks}.

In this figure, we show sample images (top row) from the CARPK \cite{lpn-carpk} dataset and its corresponding CAM generated from the simple one-look VGG-GAP model (middle row) and its enhanced version with our regulation approach (bottom row). Following the layout proposal network, we remove the last set of convolution layers from VGG. For the image in the first column, the model has a lower emphasis on a few cars under the shadow of the bridge and probably misses some of them. However, it compensates for those false negatives by indicating comparatively stronger activations on the top of the trains, which have similar brightness and texture to white cars in the image. Also for the image in the second column, the simple network demonstrates high activations in several spots in the painted side-wall (especially in $4$ car-like textured regions). These types of false detections are hard to avoid with the minimal ground-truth information of scalar object counts because the network only knows that it must learn to find a number of similar sub-regions over the whole training dataset, irrespective of the fact that the detected region truly belongs to any of the true objects or not. Consequently, this overly simple learning objective suffers from poor generalization capability. For example, if the trains and the side-walls are inpainted with neighboring road textures in the images in Figure \ref{fig:drawbacks} and fed to the network as test samples, the network will underestimate the total count even though the new images are similar to their original versions, except for minimal background alteration.

\begin{figure}[t]
\centering
\includegraphics[width=\textwidth]{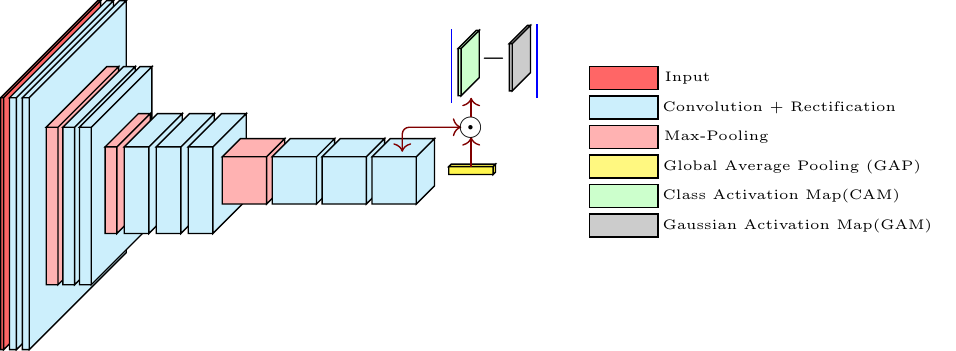}
\caption{Deep architecture used in this paper. As the baseline model, we take the first 4 sets of convolution and nonlinear activation layers from VGG16 \cite{vgg} network and attach with it a global average pooling (GAP) \cite{nin} and a linear layer to generate the scalar count. To include HR while training, we generate CAM from the last convolution layer and linear layer and compare it against the approximation of the ground-truth gaussian activation map (GAM). Note the double arrow (dark red) on top of the last convolution layer indicates that we back-propagate the CAM-GAM errors only in the convolution layers, not in the linear layer in the backend of the network. Also, we replace ReLU \cite{alexnet} with its parametric version \cite{prelu}}
\label{fig:arch}
\end{figure}

\begin{figure}[t]
\centering
\includegraphics[width=0.325\textwidth]{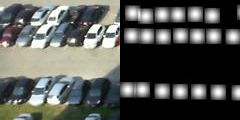}
\includegraphics[width=0.325\textwidth]{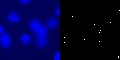}
\includegraphics[width=0.325\textwidth]{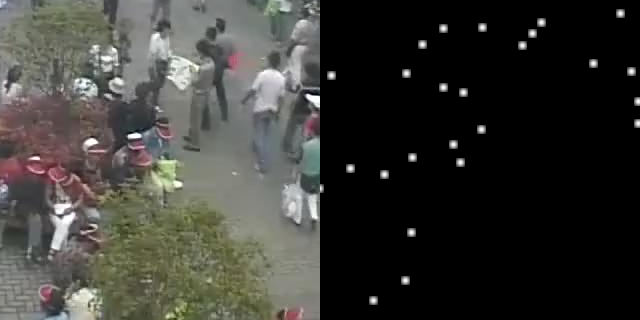}
\caption{Sample cropped images (left-side) paired with the corresponding ground-truth Gaussian activation maps (GAM) generated from the dot annotations (right-side) for CARPK \cite{lpn-carpk} (left), VGG-Cells \cite{zisserman-nips-count} (middle), and WorldExpo \cite{cross-scene-sjtu} (right) datasets.}
\label{fig:gas_gt}
\end{figure}

Our goal is to provide localization information with as light weight labeling as possible. We have devised a strategy of using simple dot annotations approximately on the center of the object, which requires similar effort to labeling the overall count, since a natural human annotation strategy could be to point at objects as they are counted.
%^
We extend the idea of class activation maps (CAM) \cite{cam-mit}, which is generated by multiplying the activations of the final convolution layer with the weights of the linear layer following a global average pooling (GAP) \cite{nin} operation and summing up the weighted responses spatially. During the training phase, we take the network generated CAM as the coarse saliency map of the image and compare it against the probable Gaussian activation map (GAM) (Figure \ref{fig:gas_gt}). The GAM is generated as an approximation of the ground-truth activation from the dot annotations using a Gaussian kernel with predefined parameters centered on the dots. We propagate the CAM-GAM errors backward along with the scalar counting error. We call this strategy heatmap regulation and the overall architecture is depicted in Figure \ref{fig:arch}. Note that, following the LPN paper \cite{lpn-carpk}, we use the first 4 sets of convolution layers of the pretrained VGG16\cite{vgg} network as the convolutional front-end followed by a global average pooling (GAP) and linear layer. Also, we back-propagate the normalized CAM errors in the convolution layers only, which is indicated by the double arrow in Figure \ref{fig:arch}. Smooth-$L_1$ \cite{fast-rcnn} and $L_1$ loss functions are used to compute CAM and count errors, respectively.

Intuitively, our heatmap regulation strategy is also useful for eliminating the need for nonlinear global information aggregation for counting via multiple fully connected (FC) layers in the backend of the network. For the counting models, where the objects under consideration, like cars, crowd, or cells occupy only a small portion of the whole image, the amalgamation of global information using FC layers should not be necessary at all. However, fully connected layers might still give better performance for less compact activation maps. Note that, for classification models, typically a gross but somewhat stronger impression about the target object evolved from the vector generated by GAP operation is sufficient to classify the objects in the images with high accuracy. On the other hand, for one-look counting models, comparatively precise activations in the object regions are needed to make the network settle for an accurate count from the vector resulting from the GAP operation at the end of the network.
In this case, the learning model might still be able to estimate the target counts with high accuracy from a less compact activation by means of aggregation of global information via the FC layers as reported in \cite{deepwheat}. As already shown in Figure \ref{fig:drawbacks}, the activations generated by the baseline one-look models are coarse and distributed more or less over the image-grid space than those generated by the corresponding HR enhanced model. This dispersed nature of the activation maps might necessitate the inclusion of FC layers for simple one-look models which is not the case for the comparatively compact and concentrated CAM we get with HR. Therefore, we expect that the HR strategy would reduce the need for global decision making via the fully connected layers by enforcing the compactness of the activation maps in the semi-global regions for individual objects, which in turn, should improve the overall performance of the models without FC layers and also prevent overfitting caused by FC layers as a by-product.

%\begin{figure}[t]
%\centering
%\includegraphics[width=0.49\textwidth]{20160331_NTU_00055.png}
%\includegraphics[width=0.49\textwidth]{20160331_NTU_00055_gas_gt.png} \\
%\includegraphics[width=0.24\textwidth]{37.png}
%\includegraphics[width=0.24\textwidth]{37_gas_gt.png}
%\includegraphics[width=0.24\textwidth]{96.png}
%\includegraphics[width=0.24\textwidth]{96_gas_gt.png} \\
%\includegraphics[width=0.49\textwidth]{108.png}
%\includegraphics[width=0.49\textwidth]{108_gas_gt.png} \\
%\caption{Sample ground-truth Gaussian activation maps (GAM) generated from the dot annotations along with corresponding images. Top, middle, and bottom rows contain samples from CARPK \cite{lpn-carpk}, VGG-Cells \cite{zisserman-nips-count}, and WorldExpo \cite{cross-scene-sjtu} datasets, respectively.}
%\label{fig:gas_gt}
%\end{figure}

\section{Experiments}

In this section, we provide the experimental results in the form of numerical performance metrics and heatmap visualizations of CAM from both the simple and HR-enhanced models.
Our models are implemented in PyTorch \cite{pytorch}. Codes and pre-trained models are publicly available here. \footnote{\textcolor{green!50!black}{\texttt{https://github.com/littleaich/heatmap{-}regulation}}}%\footnote{\url{https://github.com/littleaich/heatmap-regulation}}
We use the following set of metrics, consistent with previous works:

\begin{equation}
\begin{cases}
a_i, t_i = \text{actual and target counts for } i^{th} sample \\
N = \text{number of samples} \\
\textit{Mean Absolute Error (MAE)} = \frac{\sum_{i}|a_i-t_i|}{N} \\
\textit{Root-Mean-Square Error (RMSE)} = \sqrt{\frac{\sum_{i}\left(a_i-t_i\right)^{2}}{N} }\\
\textit{\%Underestimate(\%U)} = \frac{\sum_{i}|a_i-t_i|I_{[a_i-t_i<0]}}{\sum_{i}t_i} \times 100\\
\textit{\%Overestimate(\%O)} = \frac{\sum_{i}|a_i-t_i|I_{[a_i-t_i>0]}}{\sum_{i}t_i} \times 100\\
\textit{\%Difference(\%D)} = \%U + \%O
\end{cases}
\label{eq:metrics}
\end{equation}

% =============== old separate figures ==============================
%\begin{figure}[]
%\centering
%\includegraphics[scale=0.26]{20161225_TPZ_00043.png}
%\includegraphics[scale=0.26]{20161225_TPZ_00131.png} \\
%\includegraphics[scale=0.26]{20161225_TPZ_00043_simple.png}
%\includegraphics[scale=0.26]{20161225_TPZ_00131_simple.png} \\
%\includegraphics[scale=0.26]{20161225_TPZ_00043_gas.png}
%\includegraphics[scale=0.26]{20161225_TPZ_00131_gas.png}
%\caption{Top row shows sample images from CARPK datasets. Middle and Bottom rows comprise heatmap of CAMs generated by baseline one-look model and HR model, respectively}
%\label{fig:carpk}
%\end{figure}

%\begin{figure}[t]
%\centering
%\includegraphics[scale=0.26]{668_Rainy.jpg}
%\includegraphics[scale=0.26]{670_Rainy.jpg} \\
%\includegraphics[scale=0.26]{668_Rainy_simple.png}
%\includegraphics[scale=0.26]{670_Rainy_simple.png} \\
%\includegraphics[scale=0.26]{668_Rainy_gas.png}
%\includegraphics[scale=0.26]{670_Rainy_gas.png}
%\caption{Sample}
%\label{fig:pucpr}
%\end{figure}

% combined new figure

\begin{figure}[t]
\centering
\includegraphics[scale=0.26]{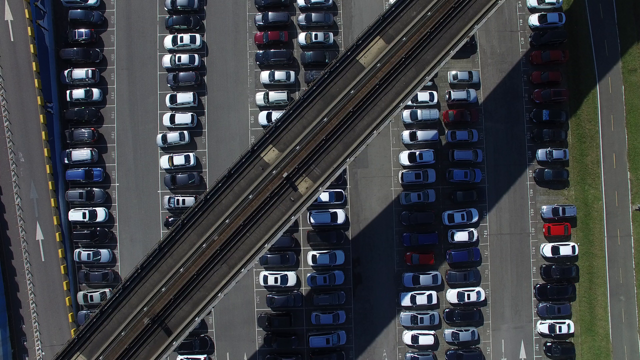}
\includegraphics[scale=0.26]{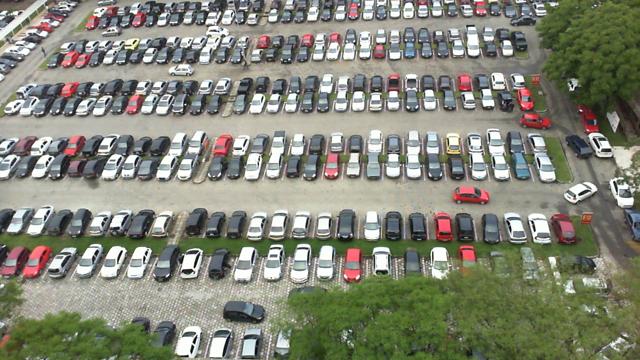} \\
\includegraphics[scale=0.26]{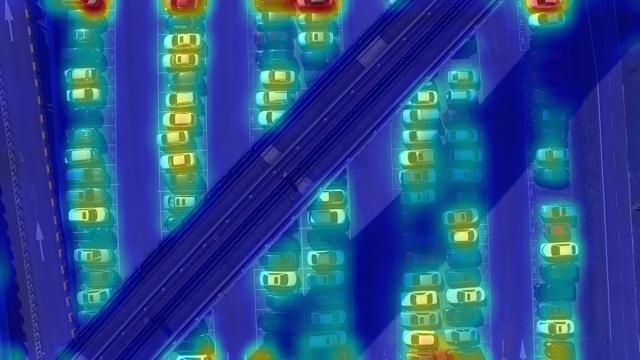}
\includegraphics[scale=0.26]{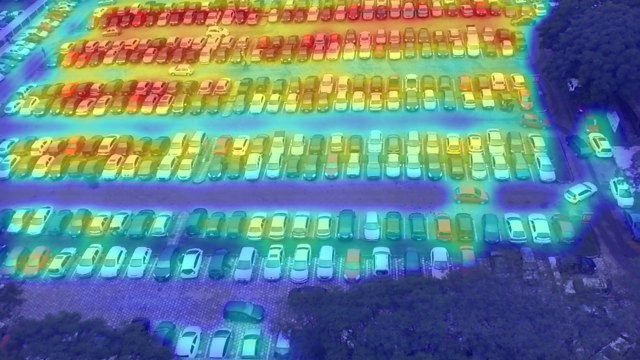} \\
\includegraphics[scale=0.26]{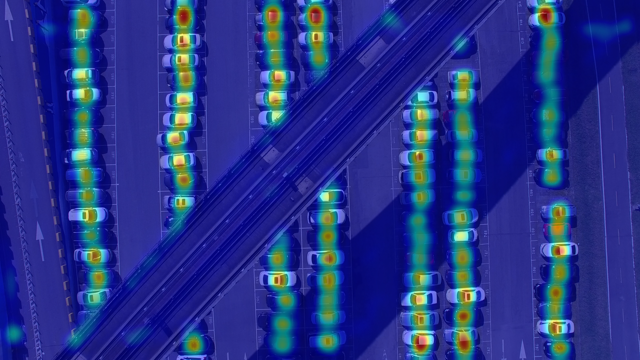}
\includegraphics[scale=0.26]{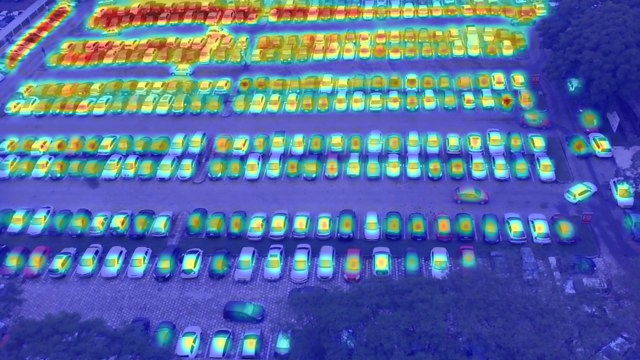}
\caption{Sample images (top row) from CARPK (left) and PUCPR+ (right) datasets, along with superimposed CAM heatmaps generated by the VGG-GAP baseline (middle row) and VGG-GAP-HR (bottom row) models.
%Activations generated by the HR model are more compact and accurate as compared to those from baseline model.
}
\label{fig:carpk-pucpr}
\end{figure}

\subsection{CARPK and PUCPR+ datasets}

%\subsubsection{CARPK:}
The CARPK dataset \cite{lpn-carpk} is reportedly the first large-scale aerial dataset for counting cars in parking lots. It contains images with a top-down view covering 4 different parking lots (Figure \ref{fig:carpk-pucpr}, Left). It includes 989 and 459 training and test samples, respectively, each of resolution $720\times 1280$. The total number of car instances in the training dataset is 42274 in the range $[1,87]$ and in the test dataset is 47500 in the range $[2,188]$. We split $10\%$ of the training data (99 images) into the validation set and use rest of the samples for training. Because of the small number of training images, we define a single epoch as 10 passes over the training images and train both the baseline and the HR enhanced version for 100 epochs each. After that, based on the performance on the validation set, we select the models after particular epochs to evaluate on the test set.

%\subsubsection{PUCPR+:}
The PUCPR+ dataset \cite{lpn-carpk} is published in the same paper as the CARPK dataset. It is a subset of PUCPR dataset \cite{pklot} which contains images covering a single parking lot with $331$ parking spaces of the same resolution ($720\times 1280$) as CARPK dataset. The images are captured using a fixed camera from a height of the $10^{th}$ floor of a building which provides a slanted view of the parking lot (Figure \ref{fig:carpk-pucpr}, Right). This dataset has in total 100 and 25 training and test samples, respectively. The total number of car instances in the training dataset is 12995 in the range $[0,331]$ and in the test dataset is 3920 in the range $[1,328]$. Like the experimental setup for CARPK dataset, we use $10\%$ of the training data (10 images) for validation. Also, we specify 100 passes over the training samples as a single epoch and run both of our models for 100 epochs.

\begin{table}[t]
\centering
\caption{Results on the CARPK test set (459 images and 47500 total counts).}
\label{tab:carpk}
\begin{adjustbox}{max width=\textwidth,center}
\begin{tabular}{|l|c|c|c|c|c|c|}
\hline
\multicolumn{1}{|c|}{Method} & \#Proposals & MAE & RMSE & \%O & \%U & \%D \\ \hline
YOLO \cite{yolo,lpn-carpk} &  & 48.89 & 57.55 & - & - & - \\ \hline
Faster R-CNN \cite{faster-rcnn,lpn-carpk} & 200 & 47.45 & 57.39 & - & - & - \\ \hline
One-Look Regression \cite{cowc,lpn-carpk} &  & 59.46 & 66.84 & - & - & - \\ \hline
LPN \cite{lpn-carpk} & 200 & 23.80 & 36.79 & - & - & - \\ \hline
LPN \cite{lpn-carpk} & 1000 & 13.72 & 21.77 & - & - & - \\ \hline
Our baseline (VGG-GAP) &  & \textbf{10.33} & \textbf{12.89} & \textbf{1.56} & \textbf{8.41} & \textbf{9.98} \\ \hline
VGG-GAP-HR &  & \textbf{7.88} & \textbf{9.30} & \textbf{0.71} & \textbf{6.91} & \textbf{7.62} \\ \hline
\end{tabular}
\end{adjustbox}
\end{table}

\begin{table}[t]
\centering
\caption{Results on the PUCPR+ test set (25 images and 3920 total counts).}
\label{tab:pucpr}
\begin{adjustbox}{max width=\textwidth,center}
\begin{tabular}{|l|c|c|c|c|c|c|}
\hline
\multicolumn{1}{|c|}{Method} & \#Proposals & MAE & RMSE & \%O & \%U & \%D \\ \hline
YOLO \cite{yolo,lpn-carpk} &  & 156.00 & 200.42 & - & - & - \\ \hline
Faster R-CNN \cite{faster-rcnn,lpn-carpk} & 400 & 39.88 & 47.67 & - & - & - \\ \hline
One-Look Regression \cite{cowc,lpn-carpk} &  & 21.88 & 36.73 & - & - & - \\ \hline
LPN \cite{lpn-carpk} & 400 & 22.76 & 34.46 & - & - & - \\ \hline
LPN \cite{lpn-carpk} & 1000 & 8.04 & 12.06 & - & - & - \\ \hline
Our baseline (VGG-GAP) &  & \textbf{8.24} & \textbf{11.38} & \textbf{0.31} & \textbf{4.95} & \textbf{5.26} \\ \hline
VGG-GAP-HR &  & \textbf{5.24} & \textbf{6.67} & \textbf{2.73} & \textbf{0.61} & \textbf{3.34} \\ \hline
\end{tabular}
\end{adjustbox}
\end{table}

For both of the car datasets, ADAM optimizer is used with initial learning rate and weight decay both equal to $0.0001$. We drop the learning rate to $10\%$ of the initial learning rate after 10 epochs and trained both models with that parameter setting for the rest of the epochs. We find similar performance using both half-sampled and full resolution images. Therefore, considering the computational complexity, we report here the performance of our models with the downsampled versions.

Both models achieve state-of-the-art performance and the HR model improves upon the baseline (Table \ref{tab:carpk} and \ref{tab:pucpr}). We were surprised that our simple VGG-GAP baseline (without any extension) performs better than more sophisticated models like LPN \cite{lpn-carpk} and one-look versions of the large ResNet and Res-ception networks \cite{cowc}. We expect the success is due to some simplifying aspects of the car counting dataset, including consistent car sizes, lack of object deformation and lack of intra-object occlusion. However, the dataset does have challenges, such as illumination variance in different parts of the image, occlusion caused by trees and flyovers, and background sub-regions or other vehicles with similar spatial statistics. Considering all these facts, a simple deep network with few stacked up convolution and nonlinear layers should be sufficient to extract most of the distinctive object features from the image.

% ian -- don't know what this means:
% which is also justified by the performance of our simple baseline model compared to other models with delicate connectivities among the layers reported in the literature.

\begin{figure}[t]
\centering
\includegraphics[width=0.325\textwidth]{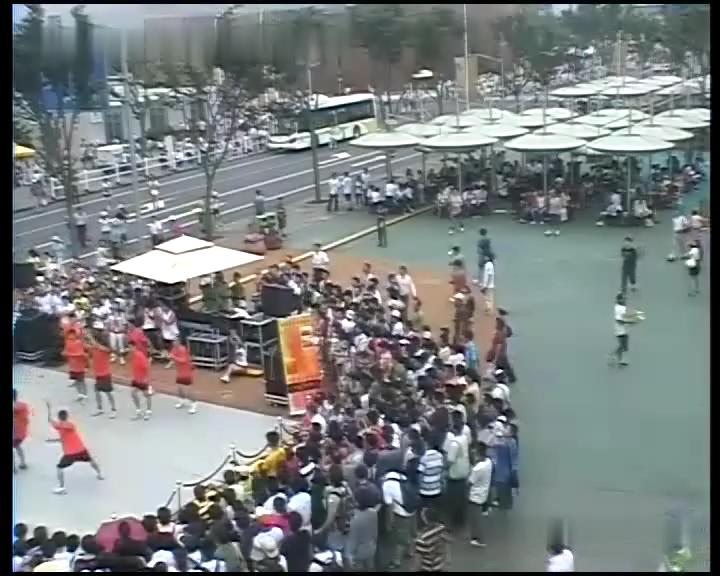}
\includegraphics[width=0.325\textwidth]{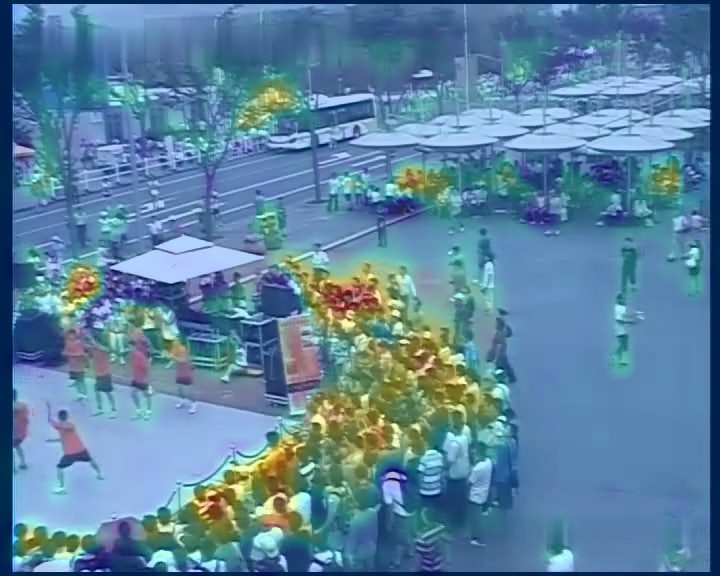}
\includegraphics[width=0.325\textwidth]{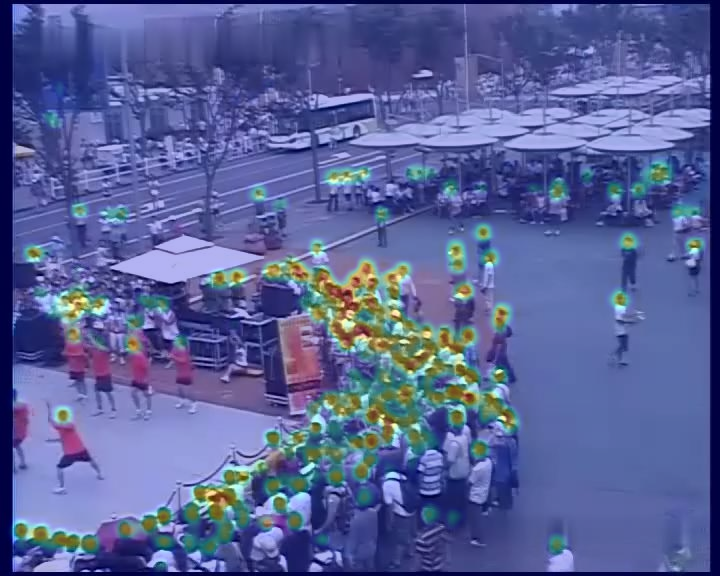} \\
\includegraphics[width=0.325\textwidth]{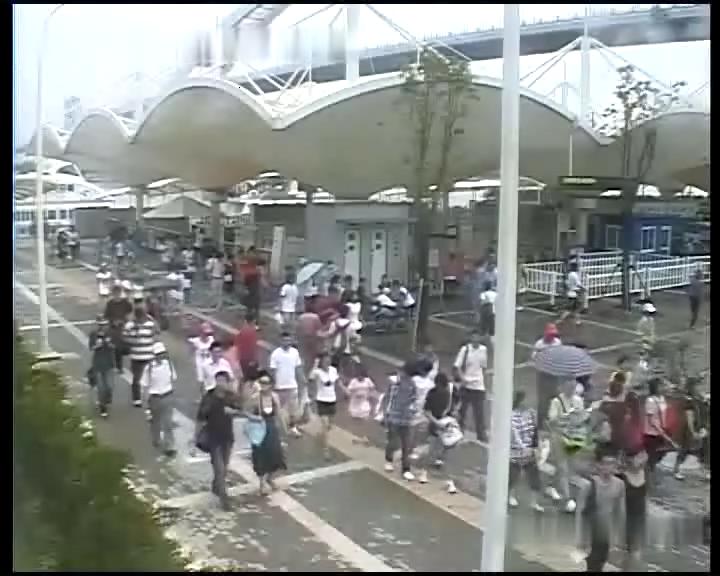}
\includegraphics[width=0.325\textwidth]{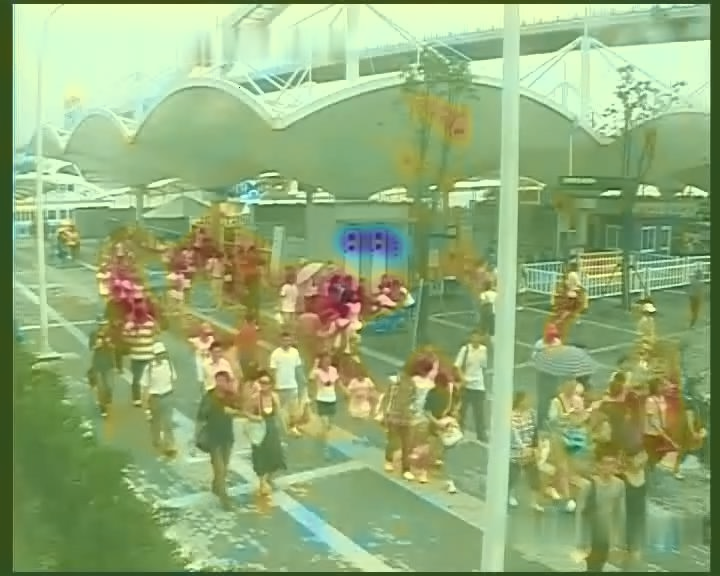}
\includegraphics[width=0.325\textwidth]{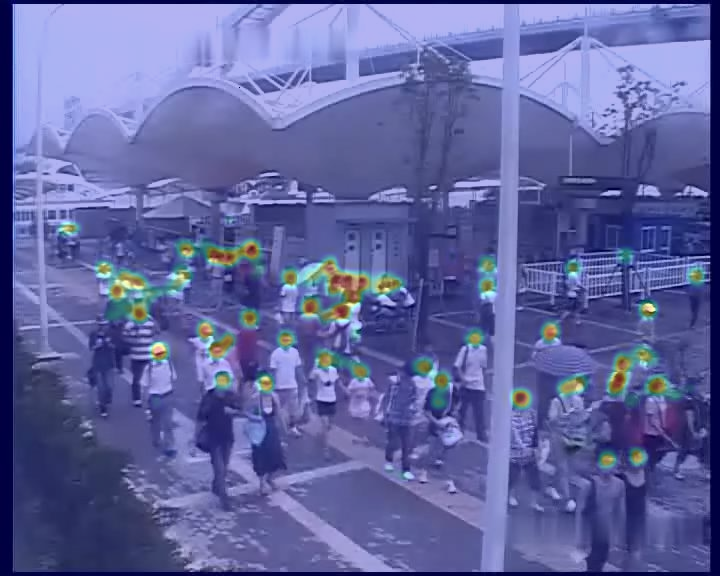}
\caption{Sample images from the WorldExpo dataset (left) with superimposed CAM heatmaps for the VGG-GAP baseline (middle) and VGG-GAP-HR (right) models.
% Note the compactness of CAM from the last column compared to the middle ones.
 }
\label{fig:crowd}
\end{figure}

Comparing CAM heatmaps for baseline (Figure \ref{fig:carpk-pucpr}, middle row) and HR (Figure \ref{fig:carpk-pucpr}, bottom row), it is evident that the simple VGG-GAP model still places significant emphasis on true object regions. However, the distribution of hotspots in the heatmap for the baseline model is widely spread, whereas the hotspots are more compactly localized by the VGG-GAP-HR model. In other words, the weight distribution in the heatmap of CAM generated by the HR model is closer to the ideal one. This observation is also reflected by the better performance of the HR model (Table \ref{tab:carpk} and \ref{tab:pucpr}).

Another interesting observation from the car counting results is that the HR strategy helps in suppressing false detections more than in assisting the detection of harder true positives. This can be explained for these particular types of datasets. First, most of the false positives are in isolated sub-regions, like train-ends, wall paintings, shadows beneath the trees, etc. Therefore, while training the network with the extra supervision with Gaussian maps, it becomes comparatively easier for the network to suppress these isolated false detections by extracting necessary contextual information. On the other hand, harder true positives are mostly located alongside easier instances. The difficulty comes with low activation in the baseline model due to partial visibility in the border regions, occlusions or different regional statistics, and some of these are confusing even for human vision. Our HR approach succeeds for some of them, such as boosting activations for black cars and cars in shadow, which is evident from the resulting evaluation metrics.

A practical observation we found is that the training error goes down faster for baseline model than for the HR model, which is consistent with the intuition behind HR. Both the simple and enhanced networks figure out most of the salient regions easily after few passes over the training set, whereas the difference lies in finding the hard cases. For harder instances, the unconstrained loss function of the simple model allows it to pick features from a reasonably arbitrary sub-region from the image, be it from the hard true positives or false background ones. However, the HR strategy applies an external force to the model to retrieve information only from the true positive regions, which slows down the convergence speed with the benefit of better generalization performance.

\begin{table}[t]
\centering
\caption{Results (MAE) on the five crowd scenes in the WorldExpo test set.}
\label{tab:crowd}
\begin{adjustbox}{max width=\textwidth,center}
\begin{tabular}{|l|c|c|c|c|c|c|}
\hline
\multicolumn{1}{|c|}{\multirow{2}{*}{Method}} & \multicolumn{5}{c|}{\begin{tabular}[c]{@{}c@{}}Test Directory Name \\ (\#Samples, \#Count)\end{tabular}} & \multicolumn{1}{l|}{\multirow{2}{*}{Average}} \\ \cline{2-6}
\multicolumn{1}{|c|}{} & \begin{tabular}[c]{@{}c@{}}104207\\ (119, 2168)\end{tabular} & \begin{tabular}[c]{@{}c@{}}200608\\ (120, 14444)\end{tabular} & \begin{tabular}[c]{@{}c@{}}200702\\ (120, 9591)\end{tabular} & \begin{tabular}[c]{@{}c@{}}202201\\ (120, 14050)\end{tabular} & \begin{tabular}[c]{@{}c@{}}500717\\ (120, 2662)\end{tabular} & \multicolumn{1}{l|}{} \\ \hline
LBP+RR \cite{cross-scene-sjtu} & 13.6 & 58.9 & 37.1 & 21.8 & 23.4 & 31.0 \\ \hline
Fiaschi et al. \cite{fiaschi2012,cross-scene-sjtu} & 2.2 & 87.3 & 22.2 & 16.4 & 5.4 & 26.7 \\ \hline
Ke et al. \cite{ke2012,cross-scene-sjtu} & 2.1 & 55.9 & 9.6 & 11.3 & 3.4 & 16.5 \\ \hline
Crowd CNN \cite{cross-scene-sjtu} & 10.0 & 15.4 & 15.3 & 25.6 & 4.1 & 14.1 \\ \hline
Fine-tuned Crowd CNN \cite{cross-scene-sjtu} & 9.8 & 14.1 & 14.3 & 22.2 & 3.7 & 12.9 \\ \hline
Crowd CNN+RR \cite{cross-scene-sjtu} & 2.0 & 29.5 & 9.7 & 9.3 & 3.1 & 10.7 \\ \hline
Our baseline (VGG-GAP) & \textbf{4.4} & \textbf{26.3} & \textbf{38.9} & \textbf{18.3} & \textbf{7.0} & \textbf{19.0} \\ \hline
VGG-GAP-HR & \textbf{3.6} & \textbf{16.8} & \textbf{24.0} & \textbf{32.6} & \textbf{6.8} & \textbf{16.8} \\ \hline
\end{tabular}
\end{adjustbox}
\end{table}

\subsection{WorldExpo dataset}
WorldExpo \cite{cross-scene-sjtu} is reportedly the largest cross-scene crowd-counting dataset, captured using 108 surveillance cameras, each viewing a different scene during the Shanghai 2010 WorldExpo. It has in total 3380 and 599 training and test samples, respectively, each of resolution $576\times 720$. The training and test sets comprise scenes from 103 and 5 different scenes, respectively. In total, there are 182301 person instances in the range $[0,334]$ in the training set and 42915 instances in the range $[1,262]$ in the test directories. We randomly choose $3\%$ of the training data (101 images) as a validation set. Like the training on car datasets, we define a single epoch as 10 passes over the training images and train the models for 100 epochs.

Results demonstrate the superiority ($>2\%$ improvement) of our HR approach compared to the simple baseline (Table \ref{tab:crowd}) with better and more compact activation maps as shown in Figure \ref{fig:crowd}. The cross-scene paper \cite{cross-scene-sjtu} achieves better accuracy than our HR approach. However, the overall approach described in that paper suffers from a number of practical limitations. First, the Crowd-CNN framework is a time-costly and multi-stage process. The CNN model there takes very small patches ($72\times 72$) as input and so, the inference on larger images ($576\times 720$) incurs time-complexity. In addition, to improve the performance of the model on the test samples extracted from unseen scenes, the authors used fine-tuning or retraining the model already trained on the training scenes using the training patches with similar view angle and scale based on the gradient of the generated perspective maps and crowd density predicted by the pretrained model on the test scenes. In this regard, even though the fine-tuning process is non-parametric, their approach is not readily generalizable to other datasets as a cross-scene crowd counting approach. Finally, the best performance in that paper is achieved in a more cumbersome manner, where the authors first predict the density map over the whole image ($576\times 720$) using overlapping $72\times 72$ patches and then use that density map as features for ridge regression (RR) to estimate the crowd density.

Contrary to this heavy multi-stage pipeline, we obtain reasonable performance using a simple network with  a straightforward and fast inference scheme with only one pass over the whole image in a single step without any perspective normalization or fine-tuning for cross-scene adaptation. Also, we generate GAM only around the dot annotations for the heads of people in the image which needs less effort compared to the crowd-counting paper \cite{cross-scene-sjtu}, where the activation maps are generated using 2 different Gaussians for both head and body considering their resolution after perspective projection. Nonetheless, it would be straightforward to add our HR strategy into this complicated, multi-stage pipeline and we expect it would further improve the performance.

\begin{figure}[t]
\centering
\includegraphics[width=0.32\textwidth]{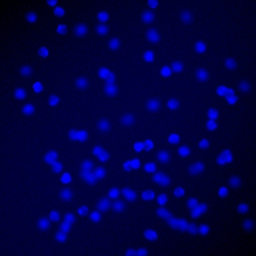}
\includegraphics[width=0.32\textwidth]{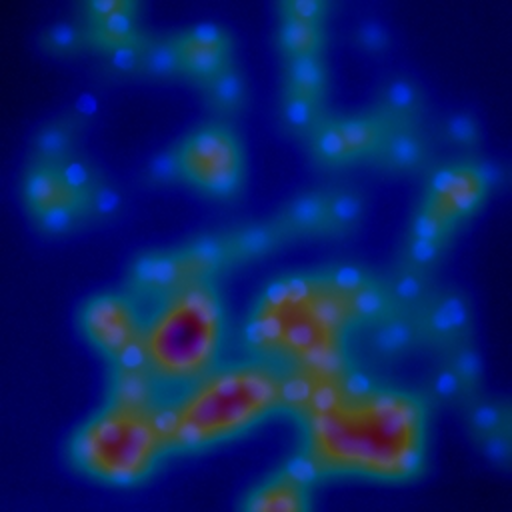}
\includegraphics[width=0.32\textwidth]{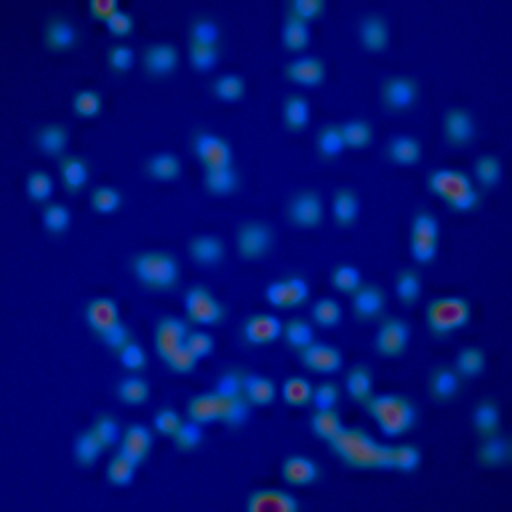} \\
\includegraphics[width=0.32\textwidth]{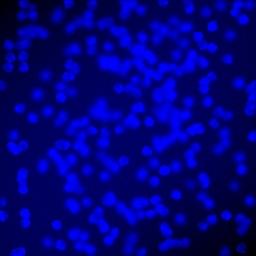}
\includegraphics[width=0.32\textwidth]{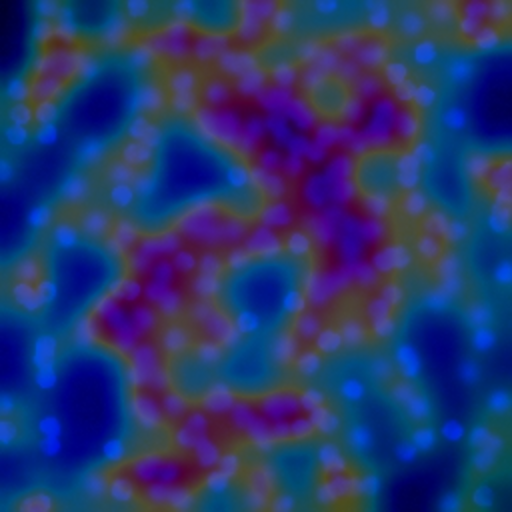}
\includegraphics[width=0.32\textwidth]{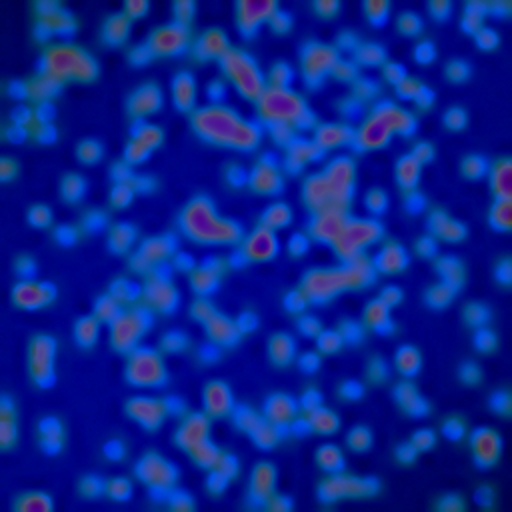}
\caption{Sample images from VGG-Cells dataset (left) with superimposed CAM heatmaps from the VGG-GAP baseline (middle) and VGG-GAP-HR (right) models.}
\label{fig:vgg-cells}
\end{figure}

\begin{table}[th]
\centering
\caption{Results on the VGG-Cells testset (100 images and 17147 total counts). N is the number of images used for training (out of 100 samples, the rest used for validation).}
\label{tab:cells}
\begin{adjustbox}{max width=\textwidth,center}
\begin{tabular}{|l|c|c|c|c|c|c|}
\hline
\multicolumn{1}{|c|}{\multirow{2}{*}{Method}} & \multicolumn{2}{c|}{MAE} & \multicolumn{2}{c|}{\%O} & \multicolumn{2}{c|}{\%U} \\ \cline{2-7}
\multicolumn{1}{|c|}{} & N=32 & N=50 & N=32 & N=50 & N=32 & N=50 \\ \hline
Lempitsky and Zisserman \cite{zisserman-nips-count} & 3.5+0.2 &  &  &  &  &  \\ \hline
Fiaschi et al. \cite{fiaschi2012} & 3.2$\pm$0.1 &  &  &  &  &  \\ \hline
Arteta et al. \cite{interactive-count} & 3.5$\pm$0.1 &  &  &  &  &  \\ \hline
Xie et al. \cite{xie2016} & 2.9$\pm$0.2 &  &  &  &  &  \\ \hline
Count-ception \cite{countception} & 2.4$\pm$0.4 & 2.3+0.4 &  &  &  &  \\ \hline
Our baseline (VGG-GAP) & \textbf{4.77} & \textbf{4.53} & \textbf{1.02} & \textbf{1.33} & \textbf{1.76} & \textbf{1.31} \\ \hline
VGG-GAP-HR & \textbf{2.95} & \textbf{2.67} & \textbf{0.76} & \textbf{0.93} & \textbf{0.96} & \textbf{0.63} \\ \hline
\end{tabular}
\end{adjustbox}
\end{table}

\subsection{VGG-Cells dataset}

VGG-Cells dataset \cite{zisserman-nips-count} comprises 200 synthetic images (100 for training and testing each) of resolution $256\times 256$, resulting from the simulation of the colonies of bacterial cells under fluorescence-light microscopy \cite{sim-cell}. The total number of cell instances in the training and test sets are 18045 (range [78, 315]) and 17147 (range [74, 317]), respectively. Following the previous benchmarks, we provide two different experimental results: ``N=32" for a 32/68 training/validation split and ``N=50" for a 50/50 training/validation split. Note that the other methods in Table \ref{tab:cells} report accuracy for multiple random trials for several random train-val splits, which we think is not appropriate for evaluating deep learning models due to their computational complexity. So, we take the first N samples (32 or 50) from the training sets for training and the rest for validation.

Again we see accuracy gains for HR over baseline (Table \ref{tab:cells}). The improvement in terms of the final activation map over the simple baseline is also evident from Figure \ref{fig:vgg-cells}. The Count-ception architecture achieves slightly better accuracy \cite{countception} than ours at the expense of more computational cost both at the training and inference stages. They estimate the redundant count maps using a fully convolutional network equipped with more sophisticated Inception-like \cite{inception-v3} modules in their architecture. This allows them to obtain multi-scale feature representations with very small input sizes $32\times 32$ to prevent overfitting, followed by redundancy elimination to get the final count. On the other hand, we obtain comparable performance by incorporating our heatmap regulation strategy into a simple VGG-GAP pipeline as before.

\section*{Acknowledgment}

This research was undertaken thanks in part to funding from the Canada First Research Excellence Fund and the Natural Sciences and Engineering Research Council (NSERC) of Canada.

\section{Conclusion}

In this paper, we propose a simple and efficient approach to improve one-look regression models for object counting using lightweight ground-truth dot annotations. Our enhancement provides near-to or better than the state-of-the-art accuracy on various counting problems with a simple VGG-GAP architecture containing only 10 stacked convolution layers. We generate ground-truth activation maps using Gaussian kernels of an approximated average size and predefined standard deviation for each experimental setup, which we find works well in practice. However, estimating these parameters in an unsupervised or a semi-supervised manner from the training dataset might lead to better performance and wider applicability, which we intend to pursue as a future direction of research.

\bibliographystyle{splncs}
\bibliography{egbib}
\end{document}